# Divergent Realities: A Comparative Analysis of Human Expert vs. Artificial Intelligence Based Generation and Evaluation of Treatment Plans in Dermatology

**Dr Dipayan Sengupta**

MD (Dermatology), Consultant dermatologist, Charnock Hospital, Kolkata

**Dr Saumya Panda**

MD (Dermatology), Professor and Head, Department of Dermatology, Jagannath Gupta Institute of Medical Sciences and Hospital, Kolkata

## Abstract

**Background:** The application of artificial intelligence (AI) in medicine is expanding from diagnostics to the complex domain of treatment planning. Evaluating the quality of AI-generated clinical plans is challenging, particularly with the emergence of advanced reasoning models whose outputs may differ from established human practice. This study aimed to compare treatment plans generated by human experts and two distinct AI models, and to investigate how these plans are assessed by both human peers and a superior AI judge.

**Methods:** Ten board-certified dermatologists, a state-of-the-art generalist AI (GPT-4o), and an advanced reasoning AI (o3) independently generated treatment plans for five complex dermatological case scenarios. The resulting 60 plans were anonymized and normalized. Evaluation was conducted in two phases: in Phase 1, the ten human experts scored the plans; in Phase 2, a superior AI model (Gemini 2.5 Pro) scored the same plans using an identical rubric.

**Results:** A profound and statistically significant "evaluator effect" was observed. In the human-led evaluation, plans generated by human experts were scored significantly higher than those from AI (mean score 7.62 vs. 7.16; p=0.0313). GPT-4o ranked 6th out of 12 participants (mean 7.383), while the advanced reasoning model, o3, ranked 11th (mean 6.974). Conversely, the AI judge's evaluation produced a complete inversion of these results, scoring AI-generated plans significantly higher than human-generated plans (mean 7.75 vs. 6.79; p=0.0313). In this phase, o3 was ranked 1st (mean 8.20), and GPT-4o was ranked 2nd (mean 7.30), with all human experts ranked below them.

**Conclusions:** The perceived quality of a clinical treatment plan is fundamentally dependent on whether the evaluator is human or AI. The advanced reasoning model, o3, judged as superior by a sophisticated AI, was poorly rated by human experts, suggesting a deep gap between data-driven algorithmic optimality and experience-based clinical heuristics. This highlights a critical challenge for the integration of AI into clinical practice. The future of effective clinical decision support likely depends not on a competition between human and AI, but on developing synergistic, human-in-the-loop systems where the rationale behind AI's novel recommendations is made transparent and explainable to the human expert.

**Introduction**

The integration of artificial intelligence (AI) into medicine is rapidly advancing, with demonstrated successes in pattern recognition tasks that have augmented diagnostic capabilities in fields such as radiology, pathology, and ophthalmology [1]. However, the frontier of AI development is moving beyond diagnostics and into the more cognitively demanding domain of clinical management. Formulating a treatment plan is a complex, iterative process that seldom has a single correct answer. It requires not only a synthesis of evidence-based knowledge but also an application of clinical judgment and experience to navigate uncertainty—a nuanced capability that is fundamentally difficult to benchmark or score objectively [2].

Within this context, dermatology presents a uniquely suitable testbed. Often narrowly viewed as a discipline where AI's primary role is the image-based detection of malignancy, dermatology in fact encompasses the complex management of chronic, systemic, and inflammatory diseases that require sophisticated, long-term therapeutic strategies [3]. The formulation of these plans involves weighing the risks and benefits of diverse treatment modalities, from topical agents to advanced biologics, and tailoring them to individual patient needs and comorbidities. This makes the speciality an ideal environment to assess the quality of AI-generated clinical reasoning beyond simple pattern recognition.

This study evaluates the treatment plans generated by human experts alongside two distinct classes of large language models. The first, GPT-4, is a state-of-the-art generalist model, widely adopted and recognized for its extensive knowledge base [4].

The second, o3, represents a more advanced architecture specifically engineered for complex, multi-step problem-solving through a more deliberative "chain-of-thought" reasoning process [5].By comparing the outputs of seasoned clinicians with these two different forms of AI, we can probe distinct facets of machine-generated clinical advice.

Given the possible distinct reasoning paradigms of clinicians and AI, this study sought to quantify the perceived quality of treatment plans across different types of authors and evaluators. The primary objective was to compare treatment plans generated by seasoned dermatologists, a generalist AI (GPT-4o), and a specialist reasoning AI (o3). Critically, we aimed to analyze these plans through two distinct evaluative lenses: that of the clinicians' peers and that of a superior AI judge [6]. This dual-phase evaluation was designed to directly investigate the potential divergence between human and algorithmic assessments of clinical quality (if any), providing insight into the future of human-AI collaboration.

## Methods

*Study Design and Participants*

This study was conducted as a comparative, observational analysis with two distinct evaluation phases. The research involved hypothetical clinical scenarios and did not include any real patient data or intervention; therefore, formal institutional review board approval was not required. Ten board-certified dermatologists practising in India, with post-residency experience ranging from 0 to 15 years, were recruited via personal invitation. Participation was voluntary, uncompensated, and all participants were aware of the study's purpose and design.

*Clinical Case Scenarios*

Five complex clinical vignettes were developed by the study authors to represent common yet challenging treatment decisions in dermatology. The cases were designed to be open-ended, lacking a single correct answer, thereby requiring nuanced clinical judgment. The scenarios included: (1) a 52-year-old male with moderate-to-severe psoriasis and multiple cardiovascular comorbidities, hesitant about injectable therapies; (2) a 29-year-old female with PCOS-related acne intending to conceive shortly; (3) a 34-year-old female with severe atopic dermatitis and recent tuberculosis exposure being considered for biologic therapy; (4) a frail, elderly woman with bullous pemphigoid and significant osteoporosis, refractory to initial steroid therapy; and (5) a 40-year-old woman with Fitzpatrick skin type V and melasma, who previously experienced post-inflammatory hyperpigmentation from a procedural intervention. The full text of each case is provided in the **Supplementary Materials (A)**.

*Treatment Plan Generation*

The five case scenarios were presented to the 10 human experts and two AI models. All participants were instructed to generate a practical, actionable treatment strategy. The AI models were accessed via the OpenAI API on April 18, 2025. The models used were gpt-4o-latest, representing the most advanced version of GPT-4o powering the ChatGPT service at that time, and o3-2025-04-16, a model engineered for complex reasoning. A single, direct prompt was used for all five cases for both models: "Create the optimum treatment strategy and next best step until next visit. For any molecule, specify the dose, duration, route, frequency, etc. Do not explain everything, just write in 2/3 sentences what the doctor should do."

*Response Normalization*

To mitigate potential biases arising from variations in writing style, length, and formatting, all 60 responses (5 cases x 12 participants) underwent a blinded, two-step normalization process. First, each raw response was processed using a separate GPT-4 API call with a detailed prompt instructing the model to standardize the language, adjust the length, and ensure stylistic similarity, while strictly preserving all core clinical information. This automated step was followed by a manual review of each normalized response by the study authors to verify that no critical information was lost or altered. Both the original and normalized versions of all treatment plans are available in the **Supplementary Materials (C)** for comparison.

*Evaluation Protocol*

The study proceeded in two evaluation phases.

*Phase 1: Human Expert Evaluation.* The 60 normalized responses were anonymized and compiled. For each case, the 10 human experts were tasked with scoring the 11 other responses (nine from their peers and two from the AI models). The evaluators were blinded to the source of each response and did not score their own submissions. The plans were presented in a randomized order for each evaluator via a custom web form accessed through a unique link. Experts were instructed to assign a single overall quality score between 0.0 and 10.0, based on a holistic assessment of five criteria: efficacy, safety, patient-centeredness, feasibility, and overall clinical appropriateness. The detailed scoring guidelines are provided in the **Supplementary Materials (B)**.

*Phase 2: Superior AI Judge Evaluation.* Following the human evaluation, all 60 normalized responses were evaluated by a superior AI judge. We utilized the gemini-2.5-pro-preview-05-06 model, accessed on June 3, 2025. The AI judge was given a persona and instructed to perform a blind assessment of each response using the exact same holistic scoring instructions and 0-10 scale that were provided to the human experts, ensuring a parallel evaluation framework.

*Statistical Analysis*

All statistical analyses were performed using Python (version 3.x), utilizing the pandas library for data manipulation, scipy for statistical tests, pingouin for intraclass correlation, and statsmodels for linear mixed-effects modeling.

## Results

The study revealed a stark, statistically significant divergence in the evaluation of treatment plans, contingent entirely on the nature of the evaluator. Human experts and the superior AI judge held diametrically opposed views on the quality of plans generated by human versus AI agents.

*Phase 1: Human Experts Favour Peer-Generated Plans*

When evaluated by the ten human dermatologists, treatment plans authored by their peers were scored significantly higher than those generated by either AI model. The 450 evaluations of human-generated plans yielded an aggregate mean score of **7.62 (SD 1.74)**, whereas the 100 evaluations of AI-generated plans produced a mean score of **7.16 (SD 1.73)**. A Wilcoxon signed-rank test confirmed that this difference was statistically significant (**p = 0.0313**).

The participant rankings reflected this preference, with human experts securing the top five positions. The AI models placed in the lower half of the rankings; **GPT-4o ranked 6th** with a mean score of 7.383 (SD 1.562), while the advanced reasoning model, **o3, ranked 11th** with a mean score of 6.974 (SD 1.836). The full ranking for all 12 participants is detailed in **Table A**.

To quantify the consistency of scoring among the human evaluators, we calculated the intraclass correlation coefficient. The analysis revealed an ICC of **0.561** (95% CI [0.385, 0.741]), indicating a moderate level of agreement among raters.

To further dissect these findings, a linear mixed-effects model was employed, controlling for variability introduced by different cases and raters. The model confirmed that the source of the plan was a significant predictor of the score. Specifically, it showed that plans generated by the o3 model were scored significantly lower than those from human experts (**p = 0.012**), even after accounting for other variables.

**Table A: Full Summary of Human Evaluation (Phase 1)**

| Rank | Plan Source | Plan Source ID | Mean | Standard deviation | count |
|---|---|---|---|---|---|
| 1 | Human | uid_ss84Ly70 | 8.172 | 1.408 | 45 |
| 2 | Human | uid_adB7pL21 | 7.838 | 1.546 | 45 |
| 3 | Human | uid_agH7mQ29 | 7.812 | 1.558 | 45 |
| 4 | Human | uid_ms91Dw04 | 7.713 | 1.379 | 45 |
| 5 | Human | uid_ag65Re18 | 7.428 | 1.904 | 45 |
| 6 | AI_GPT4o | uid_gpt4_W1X2 | 7.383 | 1.562 | 50 |
| 7 | Human | uid_ac9ZkM54 | 7.289 | 1.67 | 45 |
| 8 | Human | uid_saH3vN63 | 7.274 | 1.884 | 45 |
| 9 | Human | uid_sd52Px46 | 7.252 | 1.629 | 45 |
| 10 | Human | uid_sp42qT88 | 7.196 | 1.537 | 45 |
| 11 | AI_o3 | uid_o3_Y6Z5 | 6.974 | 1.836 | 50 |
| 12 | Human | uid_sb34UxQ9 | 6.872 | 2.27 | 45 |

*Phase 2: Superior AI Judge Favors AI-Generated Plans*

The evaluation conducted by the Gemini 2.5 Pro model produced a complete reversal of the Phase 1 findings. The AI judge scored AI-generated plans significantly higher than those created by human experts (**p = 0.0313**, Wilcoxon signed-rank test). The mean score for AI-generated plans was **7.75 (SD 1.67)**, compared to **6.79 (SD 1.74)** for human-generated plans.

This reversal was most evident in the participant rankings. The o3 model, which was ranked 11th by humans, was elevated to **1st place** by the AI judge with a mean score of **8.20 (SD 1.77)**. GPT-4o was ranked **2nd** with a mean score of **7.30 (SD 1.51)**. Consequently, all ten human experts were ranked below the two AI models. The full ranking from the Gemini evaluation is shown in **Table B**.

**Table B: Full Summary of LLM Judge Evaluation (Phase 2)**

| Rank | Plan Source | Plan Source ID | Mean | Standard Deviation | count |
|---|---|---|---|---|---|
| 1 | AI_o3 | uid_o3_Y6Z5 | 8.2 | 1.772 | 5 |
| 2 | AI_GPT4o | uid_gpt4_W1X2 | 7.3 | 1.512 | 5 |
| 3 | Human | uid_saH3vN63 | 7.22 | 1.482 | 5 |
| 4 | Human | uid_sp42qT88 | 7.16 | 1.099 | 5 |
| 5 | Human | uid_ms91Dw04 | 7.12 | 1.489 | 5 |
| 6 | Human | uid_agH7mQ29 | 7 | 1.938 | 5 |
| 7 | Human | uid_ac9ZkM54 | 6.9 | 1.821 | 5 |
| 8 | Human | uid_adB7pL21 | 6.86 | 2.116 | 5 |
| 9 | Human | uid_ag65Re18 | 6.84 | 1.808 | 5 |
| 10 | Human | uid_sd52Px46 | 6.64 | 2.032 | 5 |
| 11 | Human | uid_ss84Ly7 | 6.2 | 1.859 | 5 |

|  | 12 | Human | uid_sb34UxQ9 0 | 6 | 2.155 | 5 |

*The Evaluator Effect*

In summary, the results demonstrate a profound "evaluator effect." The preferred treatment plans were entirely dependent on the evaluator's nature. A comprehensive comparison of the dramatic shift in rankings for each participant between the two evaluation phases is presented in **Table C**. The complete, disaggregated scoring data for all 550 human evaluations and 60 AI judge evaluations are available as **Supplementary Material (D)**.

**Table C: Comparison of Rankings and Scores Across Evaluation Phases**

| Plan Source | Rank (Human Eval) | Mean Score (Human Eval) | Rank (Gemini Eval) | Mean Score (Gemini Eval) | Rank Change (Human - Gemini) |
|---|---|---|---|---|---|
| AI_o3 | 11 | 6.974 | 1 | 8.2 | +10 |
| AI_GPT4o | 6 | 7.383 | 2 | 7.3 | +4 |
| Human | 1 | 8.172 | 11 | 6.2 | -10 |
| Human | 2 | 7.838 | 8 | 6.86 | -6 |
| Human | 3 | 7.812 | 6 | 7 | -3 |
| Human | 4 | 7.713 | 5 | 7.12 | -1 |
| Human | 5 | 7.428 | 9 | 6.84 | -4 |
| Human | 7 | 7.289 | 7 | 6.9 | 0 |
| Human | 8 | 7.274 | 3 | 7.22 | +5 |
| Human | 9 | 7.252 | 10 | 6.64 | -1 |
| Human | 10 | 7.196 | 4 | 7.16 | +6 |
| Human | 12 | 6.872 | 12 | 6 | 0 |

## Discussion

The findings of this study reveal a profound and paradoxical "evaluator effect," where the perceived quality of a clinical treatment plan is fundamentally dependent on the nature of the evaluator. We observed a near-perfect inversion of rankings: human dermatology experts systematically favored plans generated by their peers, while a superior AI judge decisively preferred plans generated by AI models. Most strikingly, the advanced reasoning model, o3, was ranked last by human experts yet first by the AI judge. This discussion will interpret these findings, focusing on the cognitive and algorithmic underpinnings of this discrepancy.

A primary driver of these results likely stems from the inherent nature of human clinical judgment and the influence of cognitive biases. Human expertise is forged through clinical experience, which is often shaped by real-world constraints such as resource availability and local practice norms. We hypothesize that despite instructions to assume no constraints, our expert cohort may have unconsciously defaulted to these ingrained, experience-based heuristics which could be shaped by **confirmation bias**, where evaluators favor plans that align with their own familiar approaches, a **status quo bias**, showing resistance to novel methods outcome), survivorship bias (focusing on "successful" examples while ignoring failures), framing bias (decisions influenced by how information is presented), and premature closure (making a judgment before all relevant information is considered). [7]

Therefore, for the human experts, Plans generated by fellow humans, likely reflecting similar experiential knowledge, would feel more intuitive and correct. In contrast, an AI's knowledge is not shaped by personal experience but by a vast corpus of literature and datasets [8], which may lead to recommendations that diverge from common practice but are technically data-supported.

This leads directly to the "o3 paradox." We believe the o3 model, by leveraging the latest global literature [5], proposed treatment plans that were highly advanced, potentially involving novel molecules or unconventional strategies that are not yet widely available or practiced by our expert cohort in India. To a human expert, an unfamiliar or unconventional plan may be perceived as impractical, poorly understood, or simply incorrect, leading to a low score. The superior AI judge (Gemini 2.5 pro), however, operating on a similar data-driven logic, would recognize the evidence-based optimality of these plans and rank them highest. The discrepancy, therefore, may not reflect a failure of the o3 model, but rather a "knowledge gap" between its comprehensive, data-driven reasoning and the localized, experience-driven expertise of the human evaluators.

The single most important lesson from our findings is that neither human experts nor AI systems, in their current state, are independently sufficient for optimal clinical planning. Each brings unique, non-overlapping strengths to the table [9]. The ideal path forward appears to be a **human-in-the-loop system** that leverages the consistency and data-processing power of AI with the holistic, contextual understanding of a human clinician [10]. However, for such collaboration to be effective, our results underscore the critical importance of **Explainable AI (XAI)**[11]. If an AI proposes a correct but unconventional plan without a clear rationale, a human expert, influenced by their inherent cognitive biases, is likely to reject it [12]. AI systems must not only provide recommendations but also articulate the "why" behind them to bridge the gap in reasoning and foster trust [13]

This study has several limitations that must be acknowledged. The primary limitations are the **small sample size** of both human experts and clinical cases, which may limit the generalizability of our findings. Furthermore, the subjective, open-ended nature of the evaluation, while reflecting real-world complexity, lacks the objective precision of a single-answer task. Our study was also confined to the single specialty of dermatology. The findings may differ in other medical fields with different diagnostic and therapeutic paradigms.

In conclusion, this study highlights a fundamental divergence in the evaluative frameworks of human and artificial intelligence. Humans evaluate through a lens of personal experience and practical heuristics, while AI evaluates through a lens of data-driven, probabilistic logic The "better" plan is therefore relative to the evaluator. The future of clinical decision support lies not in determining which is superior, but in creating synergistic systems where AI augments human expertise with its vast knowledge, and humans guide AI with their irreplaceable understanding of nuance and patient context.

## Supplementary Materials

(will be available in the final published version)

   A. *Full Text of Clinical Case Scenarios*
   B. *AI Prompts and Evaluator Instructions*
   C. *Original (Pre-Normalization) Responses & after Normalization*
   D. *Complete Disaggregated Scoring Data (by human & AI evaluator)*

## Acknowledgements

We are grateful to the ten dermatology experts who voluntarily participated in this study. Their time and clinical insights were invaluable to this research. Their name will be mentioned in the final published version.

## Funding

This research received no external funding.

## Conflicts of Interest

The authors declare no conflicts of interest.

## Ethics Approval

No Ethics Approval (Not applicable)

## Declaration of generative AI and AI-assisted technologies in the writing process

During the preparation of this work, the authors used GPT-4o (ChatGPT) to improve readability and minor language formatting. After using this tool/service, the authors reviewed and edited the content as needed and take full responsibility for the content of the publication.